\def\year{2022}\relax
\documentclass[letterpaper]{article} 
\usepackage{aaai21}  
\usepackage{times}  
\usepackage{helvet} 
\usepackage{courier}  
\usepackage[hyphens]{url}  
\usepackage{graphicx} 
\usepackage{natbib}  
\usepackage{caption} 
\usepackage{bm}
\usepackage{amssymb}
\usepackage{latexsym}
\usepackage{dsfont}
\usepackage{multirow}
\usepackage{amsfonts}
\usepackage{amsmath}
\usepackage{color}

\newcommand{\tabincell}[2]{\begin{tabular}{@{}#1@{}}#2\end{tabular}}

\title{Exploring Motion and Appearance Information for Temporal Sentence Grounding}
\author{Daizong Liu\textsuperscript{\rm 1,2}, Xiaoye Qu\textsuperscript{\rm 3}, Pan Zhou\textsuperscript{\rm 1*}, Yang Liu\textsuperscript{\rm 4} \\}
\affiliations{
\textsuperscript{\rm 1}The Hubei Engineering Research Center on Big Data Security, School of
Cyber Science and Engineering, \\ Huazhong University of Science and Technology\\
\textsuperscript{\rm 2}School of Electronic Information and Communication, Huazhong University of Science and Technology\\
\textsuperscript{\rm 3}Huawei Cloud \ 
\textsuperscript{\rm 4}Harbin Institute of Technology \
\\
\{dzliu, panzhou\}@hust.edu.cn, quxiaoye@huawei.com, liu.yang@hit.edu.cn
}

\begin{document}
\maketitle
\begin{abstract}
This paper addresses temporal sentence grounding. Previous works typically solve this task by learning frame-level video features and align them with the textual information. A major limitation of these works is that they fail to distinguish ambiguous video frames with subtle appearance differences due to frame-level feature extraction. Recently, a few methods adopt Faster R-CNN to extract detailed object features in each frame to differentiate the fine-grained appearance similarities.
However, the object-level features extracted by Faster R-CNN suffer from missing motion analysis since the object detection model lacks temporal modeling.
To solve this issue, we propose a novel \textbf{M}otion-\textbf{A}ppearance \textbf{R}easoning \textbf{N}etwork (MARN), which incorporates both motion-aware and appearance-aware object features to better reason object relations for modeling the activity among successive frames.
Specifically, we first introduce two individual video encoders to embed the video into corresponding motion-oriented and appearance-aspect object representations. Then, we develop separate motion and appearance branches to learn motion-guided and appearance-guided object relations, respectively. At last, both motion and appearance information from two branches are associated to generate more representative features for final grounding.
Extensive experiments on two challenging datasets (Charades-STA and TACoS) show that our proposed MARN significantly outperforms previous state-of-the-art methods by a large margin.
\end{abstract}

\section{Introduction}
Temporal sentence grounding is an important topic of cross-modal understanding in computer vision. Given an untrimmed video, it aims to locate a segment that contains the interested activity corresponding to the sentence description. There are several related tasks proposed involving both video and language, such as video summarization \cite{song2015tvsum,chu2015video}, 
video question answering \cite{gao2019structured,le2020hierarchical},
and temporal sentence grounding \cite{gao2017tall,anne2017localizing}. 
Among them, temporal sentence grounding is the most challenging task due to its detailed multi-modal interaction and complicated context reasoning. 

\begin{figure}[t]
\centering
\includegraphics[width=0.48\textwidth]{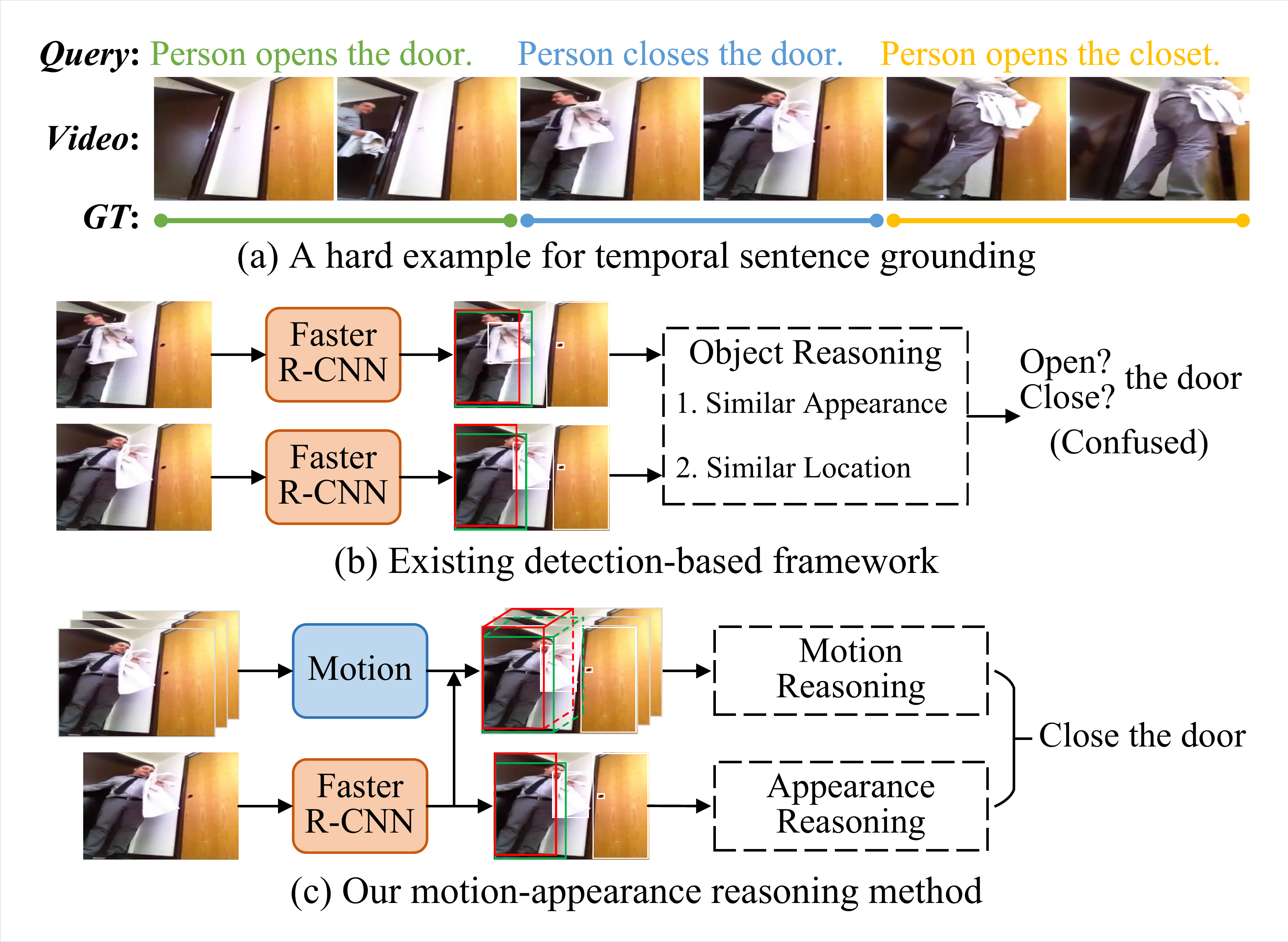}
\caption{(a) A hard example where the video contains several semantically similar segments. (b) Existing detection-based framework only extracts appearance-aware object information, and fails to distinguish similar motions ``open" and ``close". (c) Our method develops additional branch to learn motion-aware object contexts, and associates the information of two branches to reason the query.}
\label{fig:intro}
\vspace{-10pt}
\end{figure}

To localize the target segment, most previous works  \cite{anne2017localizing,gao2017tall,chen2018temporally,zhang2019cross,liu2020reasoning,liu2020jointly,zhang2019learning,liu2021adaptive,liu2021progressively} first pre-define abundant video segments as proposals, and then match them with the sentence query for ranking. The best segment proposal with the highest matching degree is finally selected as the target segment. Instead of proposal-based paradigm, some proposal-free works \cite{rodriguez2020proposal,chenrethinking,yuan2019find,mun2020local,zhang2020span} propose to directly regress the start and end timestamps of the target segment for each frame. 
Although above two kinds of works bring significant improvements in recent years, all of them extract frame-level video features to model the semantic of the target activity, which captures the redundant background information and fails to perceive the fine-grained differences among video frames with high similarity.
As shown in Figure \ref{fig:intro} (a), for the semantically similar queries ``Person opens the door" and ``Person opens the closet", modeling the temporal relations by frame-wise features can capture the same action ``open", but it is not enough to distinguish the local details of different objects (``door" and ``closet") in these frames.

Recently, detection-based approaches \cite{zeng2021multi,zhang2020object,zhang2020does} have been proposed to capture fine-grained object appearance features in each frame and achieved promising results. Among them, \cite{zeng2021multi} focus on temporal sentence grounding and learn spatio-temporal object relations to reason the semantic of target activity, while other works consider spatio-temporal object grounding task where an object rather than a video segment is retrieved.
These works can well alleviate the issue of indistinguishable local appearances, such as ``door" and ``closet", by learning the representations of objects in the frame. 
However, methods like \cite{zeng2021multi} generally extract object features by the object detection model like Faster R-CNN \cite{ren2015faster}, which lacks the object-level motion context to model the temporal action of a specific object, thus degenerating the performance on similar events. As shown in Figure 1 (b), 
it is hard for detection-based methods to distinguish the similar motions ``open" and ``close" by learning the object relations in the successive frames, since the objects extracted by Faster R-CNN have similar appearance and spatial positions in these frames. Therefore, the motion context plays an important role in modeling the consecutive states or actions for objects. \textbf{How to effectively integrate the action knowledge from motion contexts and the appearance knowledge from detection model to compose the complicated activity is an emerging issue.}

To this end, in this paper, we propose a novel \textbf{M}otion- \textbf{A}ppearance \textbf{R}easoning \textbf{N}etwork (MARN),
which incorporates motion contexts into appearance-based object features for better reasoning the semantic relations among objects.
Specifically, we detect and obtain appearance-aware object representations by a Faster R-CNN model, and simultaneously apply RoIAlign \cite{he2017mask} on the 3D feature maps from C3D network \cite{tran2015learning} for motion-aware object features extraction. Then, we develop separate branches to reason the motion-guided and appearance-guided object relations, respectively. In each branch, we interact object features with query information for query-related object semantic learning and adopt a fully-connected object graph for spatio-temporal semantic reasoning. At last, we represent frame-level features by aggregating object features inside the frame, and introduce a motion-appearance associating module to integrate representative information from two branches for final grounding.

The main contributions of this work are three-fold:
\begin{itemize}
    \item To the best of our knowledge, we are the first work that explores both motion-aware and appearance-aware object information, and proposes a novel Motion-Appearance reasoning network for temporal sentence grounding. 
    \item We devise motion and appearance branches to capture action-oriented and appearance-guided object relations. A motion-appearance associating module is further proposed to integrate the most representative features from two branches for final grounding.
    \item We conduct extensive experiments on two challenging datasets Charades-STA and TACoS. The experimental results show that our proposed MARN outperforms other state-of-the-art approaches with a large margin.
\end{itemize}

\section{Related Work}

\noindent \textbf{Temporal Sentence Grounding.}
The task of temporal sentence grounding is introduced by \cite{gao2017tall} and \cite{anne2017localizing}, which aims to identify the start and end timestamps of one specific video segment semantically corresponding to the given sentence query. Most previous works \cite{anne2017localizing,chen2018temporally,zhang2019cross,yuan2019semantic,zhang2019learning,qu2020fine,liu2022memory} localize the target segment via generating video segment proposals. They utilize sliding windows or pre-defined segment proposals to generate segment candidates, and then match them with the query. Instead of using segment candidates, some works \cite{chenrethinking,yuan2019find,mun2020local,zhang2020span,liu2022unsupervised} directly regress the start and end timestamps after interacting the whole video with query.
However, these two types of methods are all based on frame-level features to capture the semantic of video activity, which fails to capture the fine-grained difference among video frames with high similarity, especially the adjacent frames near the segment boundary.
Recently, detection-based approaches \cite{zeng2021multi,zhang2020object,zhang2020does} have been proposed to capture object appearances in the frame, which leads to more precise localization. However, they only adopt object features extracted by detection models, thus cannot obtain the motion information of each object.
As for \cite{zeng2021multi}, it builds dual textual and visual object graphs for cross-modal graph matching and directly utilize object features for grounding.
In this paper, we only construct visual graph and interact object features with textual information for semantic enhancement, then integrate object features inside each frame to represent frame-level features for grounding. Moreover, considering motion contexts play a key role \cite{seo2021attend} in modeling the consecutive states or actions, we extract both motion-aware and appearance-aware object features to capture action-oriented and appearance-aspect contexts for more accurate spatio-temporal object reasoning.



\begin{figure*}[t]
\centering
\includegraphics[width=1.0\textwidth]{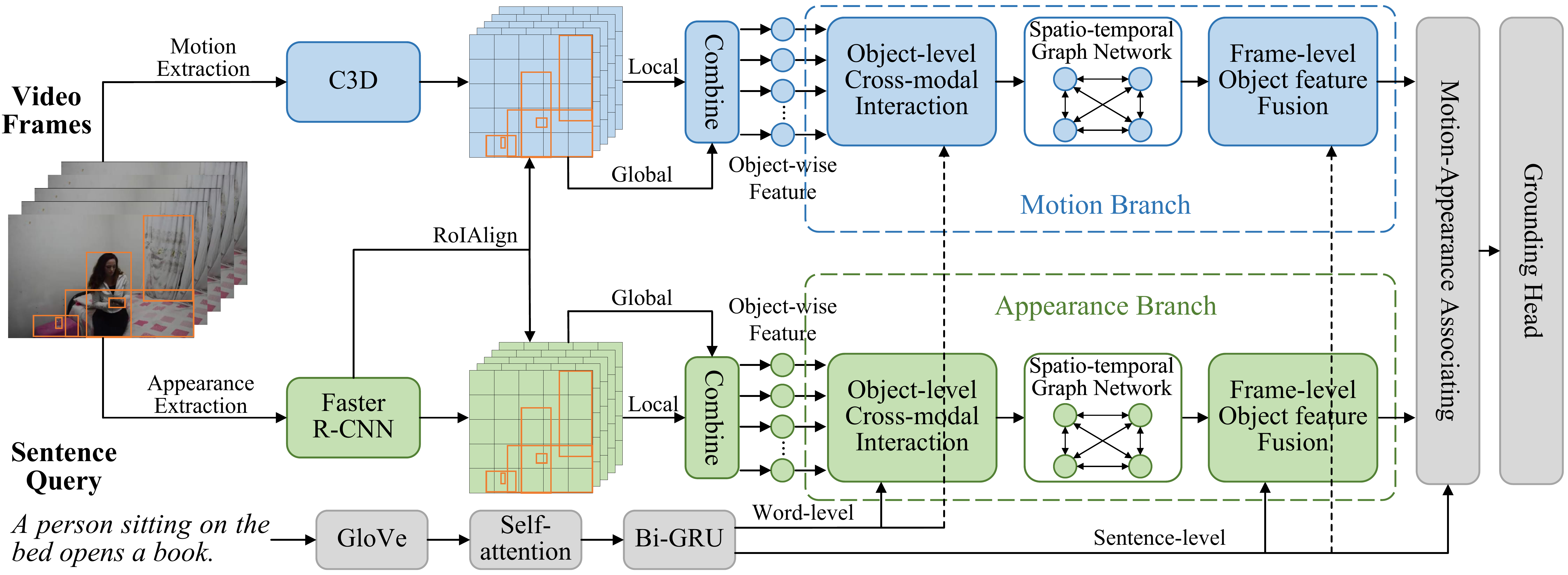}
\caption{Overall pipeline of the proposed network MARN. We first utilize video and query encoders to extract motion- and appearance-aware object features, word- and sentence-level query features. Then, we develop separate motion and appearance branches for specific cross-modal object reasoning. At last, we associate motion-appearance information for final grounding.}
\label{fig:pipeline}
\vspace{-10pt}
\end{figure*}

\section{Model}
Given an untrimmed video $V$ and a sentence query $Q$, we present the video as $V=\{v_t\}_{t=1}^T$ where $v_t$ denotes the $t$-th frame and $T$ denotes the frame number. Analogously, the sentence query is represented as $Q=\{q_n\}_{n=1}^N$ where $q_n$ denotes the $n$-th word and $N$ denotes the total number of words. The temporal sentence grounding (TSG) task aims to retrieve the video segment including both start and end timestamps that is most relevant to the sentence query.

In this section, we propose the Motion-Appearance Reasoning Network (MARN), which incorporates both motion and appearance information to reason object relations for modeling the activity among successive frames. As shown in Figure \ref{fig:pipeline}, we first extract motion- and appearance-aware object features, and then develop separate motion and appearance branches to learn fine-grained action-oriented and appearance-aspect object relations. Specifically, in each branch, after interacting object-query features to filter out irrelevant object features, we reason the relations between foremost objects with a spatio-temporal graph and represent frame-level features by fusing its contained objects features. At last, we associate motion and appearance frame-level features for final grounding.

\subsection{Video and Query Encoders}
\noindent \textbf{Video encoder.}
Unlike previous detection-based methods only using Faster R-CNN \cite{ren2015faster} pre-trained on image detection dataset to extract appearance-aware object features, we consider additionally extracting motion-aware object information to obtain action-oriented features for temporal modeling.
Specifically, \textbf{for appearance features}, 
we first sample fixed $T$ frames from the original video averagely, and then obtain $K$ objects from each frame using Faster R-CNN that is built on a ResNet \cite{he2016deep} backbone.
Therefore, there are total $T \times K$ objects in a single video, and we can represent their appearance features as $\bm{V}^a_{local}= \{\bm{o}^a_{t,k}, \bm{b}_{t,k}\}_{t=1,k=1}^{t=T,k=K}$, where $\bm{o}^a_{t,k} \in \mathbb{R}^{D}, \bm{b}_{t,k} \in \mathbb{R}^{4}$ denotes the local appearance feature and bounding-box position of the $k$-th object in $t$-th frame. 
Since the global feature of the whole frame also contains the non-local information of its internal objects, we utilize another ResNet model with a linear layer to generate frame-wise appearance representation $\bm{V}_{global}^a \in \mathbb{R}^{T \times D}$.
\textbf{For motion features}, we first extract the feature maps of each video clip from the last convolutional layer in C3D \cite{tran2015learning} network, and then apply RoIAlign \cite{he2017mask} on such feature maps and use object bounding-box locations $\bm{b}_{t,k}$ to generate motion-aware object features $\bm{V}^m_{local}= \{\bm{o}^m_{t,k}, \bm{b}_{t,k}\}_{t=1,k=1}^{t=T,k=K}$. To extract the clip-wise global features $\bm{V}_{global}^m \in \mathbb{R}^{T \times D}$, we directly apply average pooling and linear projection to the extracted feature maps of C3D.

Since it is necessary to consider each object’s both spatial and temporal locations for reasoning object-wise relations, we add a position encoding to object-level local features in both appearance and motion representations as:
\begin{equation}
    \bm{v}_{t,k}^a = \text{FC}([\bm{o}^a_{t,k};\bm{e}^b;\bm{e}^t]), \bm{v}_{t,k}^m = \text{FC}([\bm{o}^m_{t,k};\bm{e}^b;\bm{e}^t]),
\end{equation}
where $\bm{e}^b = \text{FC}(\bm{b}_{t,k})$, $\text{FC}(\cdot)$ is the fully connected layer, $\bm{e}^t$ is obtained by position encoding \cite{mun2020local} according to each frame’s index. Thus,  $\widehat{\bm{V}}_{local}^a=\{\bm{v}_{t,k}^a\}_{t=1,k=1}^{t=T,k=K}, \widehat{\bm{V}}_{local}^m=\{\bm{v}_{t,k}^m\}_{t=1,k=1}^{t=T,k=K}$. Similarly, we add position encoding into two global representations as:
\begin{equation}
   \widehat{\bm{V}}^a_{global} = \text{FC}([\bm{V}^a_{global};\textbf{e}^T]),
    \widehat{\bm{V}}^m_{global} =\text{FC}([\bm{V}^m_{global};\textbf{e}^T]).
\end{equation}
At last, we expand the above two global features from size of $T \times D$ to size of $(T \times K) \times D$, and concatenate the local object features with corresponding global features to reflect the context in objects as:
\begin{equation}
    \bm{F}^a = \text{FC}([\widehat{\bm{V}}_{local}^a;\widehat{\bm{V}}^a_{global}]),
    \bm{F}^m = \text{FC}([\widehat{\bm{V}}_{local}^m;\widehat{\bm{V}}^m_{global}]),
\end{equation}
where $\bm{F}^a=\{\bm{f}_{t,k}^a\}_{t=1,k=1}^{t=T,k=K},\bm{F}^m = \{\bm{f}_{t,k}^m\}_{t=1,k=1}^{t=T,k=K} \in \mathbb{R}^{(T \times K)\times D}$ denote the final encoded object-level features.

\noindent \textbf{Query encoder.}
We first utilize the Glove \cite{pennington2014glove} to embed each word into dense vector, and then employ multi-head self-attention \cite{vaswani2017attention} and Bi-GRU \cite{chung2014empirical} to encode its sequential information. The final word-level features can be denoted as $\bm{Q}=\{\bm{q}_n\}_{n=1}^N \in \mathbb{R}^{N \times D}$, and the sentence-level feature $\bm{q}_{global} \in \mathbb{R}^{D}$ can be obtained by concatenating the last hidden unit outputs in Bi-GRU.

\subsection{Cross-modal Object Reasoning}
After extracting appearance- and motion-aware object representations,
we develop two separate branches to reason both motion-guided and appearance-guided object relations with cross-modal interaction. In each branch, we first interact object features with the query to enhance their semantic, and then reason object relations in a spatio-temporal graph. A query-guided attention module is further developed to fuse the object information within each frame to represent frame-level features.

\noindent \textbf{Cross-modal interaction.}
Learning correlations between visual features and query information is important for query-based video grounding, which helps to highlight the relevant object features corresponding to the query while weakening the irrelevant ones. Specifically, for the $k$-th object in the $t$-th frame in the motion branch, we interact its feature $\bm{f}^m_{t,k}$ with word-level query features $\{\bm{q}_n\}_{n=1}^N$ by:
\begin{equation}
    \bm{M}^m_{t,k,n} = \bm{\text{w}}^{\top} \text{tanh}(\bm{W}_1^m \bm{f}^m_{t,k} + \bm{W}_2^m \bm{q}_n + \bm{b}^m_1),
\end{equation}
where $\bm{W}_1^m,\bm{W}_2^m$ are learnable matrices, $\bm{b}^m_1$ is the bias vector and the $\bm{\text{w}}^{\top}$ is the row vector as in \cite{zhang2019cross}. The query-enhanced object features $\widehat{\bm{f}}^m_{t,k}$ can be obtained by:
\begin{equation}
    (\bm{f}^m_{t,k})' = \sum_{n=1}^N \text{softmax}(\bm{M}^m_{t,k,n}) \bm{q}_n,
\end{equation}
\begin{equation}
    \widehat{\bm{f}}_{t,k}^m= \sigma(\bm{W}_3^m (\bm{f}^m_{t,k})'+\bm{b}^m_2) \odot \bm{f}^m_{t,k},
\end{equation}
where $\sigma$ is the sigmoid function, $\odot$ represents element-wise product, $\bm{W}_3^m,\bm{b}_2^m$ are parameters. $\widehat{\bm{F}}^m=\{\widehat{\bm{f}}_{t,k}^m\}_{t=1,k=1}^{t=T,k=K} \in \mathbb{R}^{(T \times K)\times D}$, and the enhanced object features $\widehat{\bm{F}}^a$ of appearance branch can be obtained in the same way.

\noindent \textbf{Spatio-temporal graph.}
Since the detected objects have both spatial interactivity and temporal continuity, as shown in Figure \ref{fig:pipeline}, we construct object graph to capture spatio-temporal relations in each branch, respectively. For motion branch, we define object-wise features $\widehat{\bm{F}}^m=\{\widehat{\bm{f}}_{t,k}^m\}_{t=1,k=1}^{t=T,k=K}$ including all objects in all frames as nodes and build a fully-connected motion graph. We adopt graph convolution network (GCN) \cite{kipf2016semi} to learn the object-relation features via message propagation.
In details, we first measure the pairwise affinity between object features by:
\begin{equation}
    \bm{A}^m = \text{softmax}((\widehat{\bm{F}}^m \bm{W}^m_4)(\widehat{\bm{F}}^m \bm{W}^m_5)^{\top}),
\end{equation}
where $\bm{W}^m_4,\bm{W}^m_5$ are learnable parameters. $\bm{A}^m \in \mathbb{R}^{(T \times K) \times (T \times K)}$ is obtained by calculating the affinity edge of each pair of objects. Two objects with strong semantic relationships will be highly correlated and have an edge with high affinity score in $\bm{A}^m$. Then, we apply single-layer GCN with residual connections to perform semantic reasoning:
\begin{equation}
    \widetilde{\bm{F}}^m = (\bm{A}^m \widehat{\bm{F}}^m \bm{W}^m_6) \bm{W}^m_7 + \widehat{\bm{F}}^m,
\end{equation}
where $\bm{W}^m_6$ is the weight matrix of the GCN layer, $\bm{W}^m_7$ is the weight matrix of residual structure. 
The output $\widetilde{\bm{F}}^m=\{\widetilde{\bm{f}}_{t,k}^m\}_{t=1,k=1}^{t=T,k=K} \in \mathbb{R}^{(T \times K)\times D}$ is the updated features for motion-aware objects. Feature $\widetilde{\bm{F}}^a$ for appearance-aware objects can be obtained in the same way.

\noindent \textbf{Object feature fusion.}
After obtaining the object features, we aim to integrate object features within each frame to represent fine-grained frame-level information under the guidance of query information. To this end, we compute the cosine similarity between object feature $\widetilde{\bm{f}}^m_{t,k}$ and the sentence-level query feature $\bm{q}_{global}$ at frame $t$ as :
\begin{equation}
    c^m_{t,k} = \frac{(\widetilde{\bm{f}}^m_{t,k})(\bm{q}_{global}\bm{W}_q)^{\top}}{||\widetilde{\bm{f}}^m_{t,k}||_2 ||\bm{q}_{global}\bm{W}_q||_2},
\end{equation}
where $\bm{W}_q$ is the linear parameter, $c^m_{t,k}$ indicates the relational score between the visual object and the given query. Then, we integrate the object features within each frame $t$ to represent query-specific frame-level feature as:
\begin{equation}
    \bm{h}^m_t = \sum_{k=1}^K \text{softmax}(c^m_{t,k}) \widetilde{\bm{f}}^m_{t,k}.
\end{equation}
The final query-specific frame-level features in both appearance and motion branches can be denoted as $\bm{H}^a=\{\bm{h}^a_t\}_{t=1}^T,\bm{H}^m=\{\bm{h}^m_t\}_{t=1}^T \in \mathbb{R}^{T \times D}$.

\begin{figure}[t]
\centering
\includegraphics[width=0.44\textwidth]{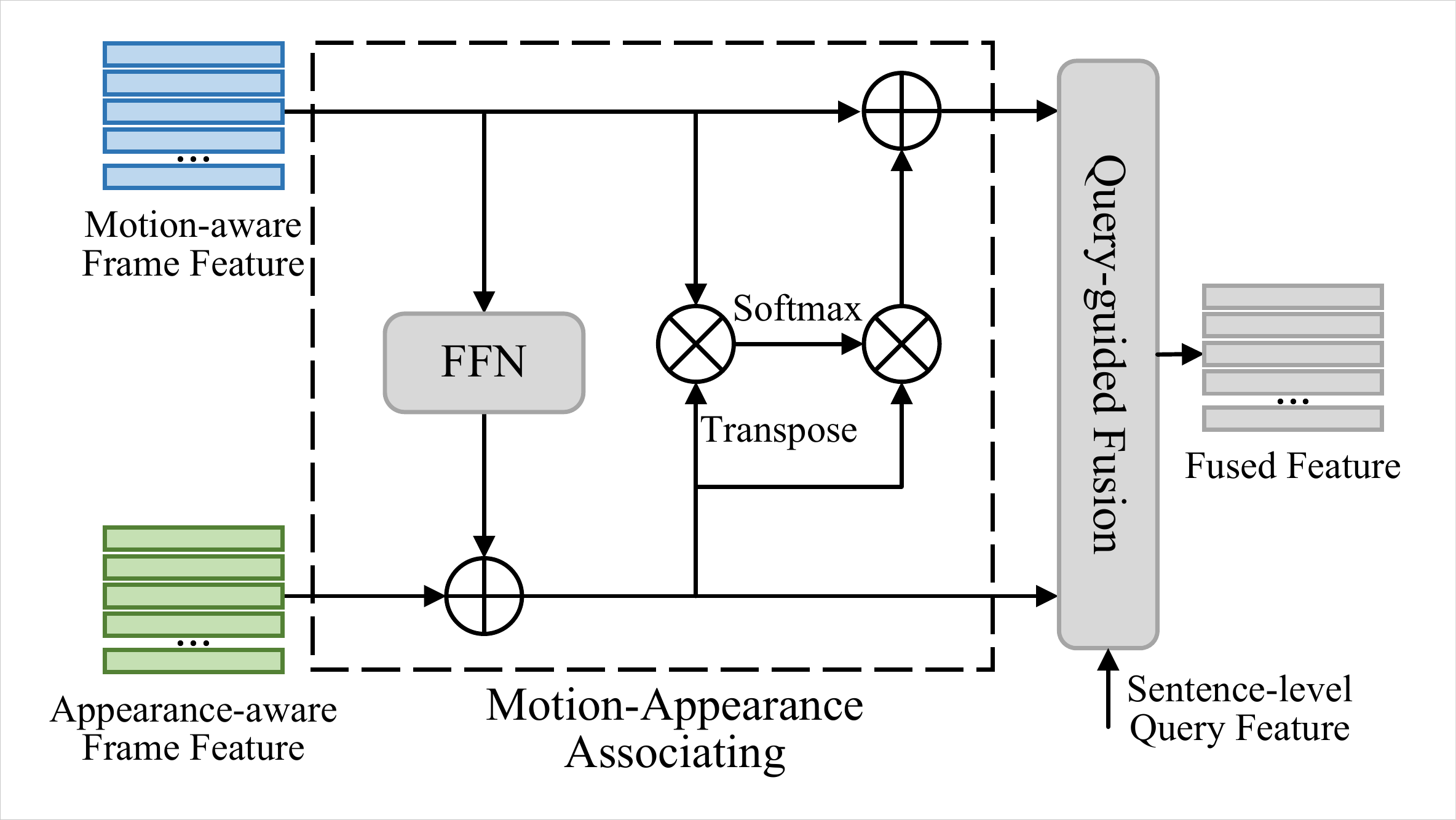}
\vspace{-8pt}
\caption{Illustration of the Motion-Appearance Associating (MAA) module, where FFN consists of three linear layers.}
\label{fig:mas}
\vspace{-10pt}
\end{figure}

\subsection{Associating Motion and Appearance}
After generating appearance- and motion-aware frame-level features $\bm{H}^a$ and $\bm{H}^m$, we develop a motion-appearance associating (MAA) module to associate their features and decide which features are most discriminative for final grounding. Module details are shown in Figure \ref{fig:mas}.

\noindent \textbf{Motion-guided appearance enhancement.}
Compared to appearance-aware feature $\bm{H}^a$, the motion-aware feature $\bm{H}^m$ captures the temporal contexts of objects. Considering the motion context contains implicit appearance information,  
we abstract the motion context and integrate it into the appearance features for generating motion-enhanced appearance features. 
Firstly, the motion feature $\bm{H}^m$ is adapted to appearance feature space through a frame-independent feed-forward network (FFN) which consists of three linear layers. Then, We add the adapted motion information to the appearance feature $\bm{H}^a$ in an element-wise way, leading to motion-enhanced appearance feature $\widehat{\bm{H}}^a$. This procedure can be formulated as:
\begin{equation}
    \widehat{\bm{H}}^a = \bm{H}^a + \text{FFN}(\bm{H}^m).
\end{equation}
In this soft and learnable way, contexts appeared in the motion space are aggregated into the appearance feature space.

\noindent \textbf{Appearance-fused motion enhancement.}
Compared to motion-aware feature $\bm{H}^m$, the feature $\bm{H}^a$ represents more on the appearance details and spatial location clues of a certain object. 
We utilize the dot-product attention to attend appearance-aware object features into motion space for inferring motion contexts as:
\begin{equation}
    \widehat{\bm{H}}^m = \bm{H}^m + \text{softmax}(\bm{H}^m (\widehat{\bm{H}}^a)^{\top})\widehat{\bm{H}}^a,
\end{equation}
where $\widehat{\bm{H}}^m$ is the appearance-enhanced motion feature.

\noindent \textbf{Query-guided motion-appearance fusion.}
To decide which information are most discriminative among appearance and motion features $\widehat{\bm{H}}^a,\widehat{\bm{H}}^m$ corresponding to the query, we integrate them under the guidance of sentence-level query features
through an attention-based weighted summation as:
\begin{equation}
\begin{split}
    \widetilde{\bm{H}} = \ & \text{softmax}(\widehat{\bm{H}}^a(\bm{q}_{global})^{\top}) \odot \widehat{\bm{H}}^a \\
    & + \text{softmax}(\widehat{\bm{H}}^m(\bm{q}_{global})^{\top}) \odot \widehat{\bm{H}}^m,
\end{split}
\end{equation}
where $\widetilde{\bm{H}} = \{\widetilde{\bm{h}}_t \}_{t=1}^T \in \mathbb{R}^{T \times D}$ is the integrated frame-level features to be fed into the last grounding head.

\subsection{Grounding Head}
With the fine-grained query-specific video features $\widetilde{\bm{H}}$, many grounding heads are plug and play. In this paper, we follow previous works \cite{zhang2019cross,yuan2019semantic} to retrieve the target video segment with pre-defined segment proposals. In details, we first define multi-size candidate segments on each frame $t$, and adopt multiple fully-connected layers to process frame-wise features $\widetilde{\bm{h}}_t$ to produce the confidence scores and temporal offsets of the segment proposals. Suppose there are $R$ proposals within the video, for each proposal whose start and end timestamp is $(\tau_s,\tau_e)$, we denote its confidence score and offsets as $o$ and $(\delta_s,\delta_e)$, where $s,e$ means the start and end. Therefore, the predicted segments of each proposal can be presented as $(\tau_s+\delta_s,\tau_e+\delta_e)$. During training, we compute the Intersection over Union (IoU) score $o^{gt}$ between each pre-defined segment proposal and the ground truth, and utilize it to supervise the confidence score as:
\begin{equation}
    \mathcal{L}_{iou} = - \frac{1}{R} \sum o^{gt} \text{log}(o)+(1-o^{gt})\text{log}(1-o).
\end{equation}
As the boundaries of pre-defined segment proposals are relatively coarse, we further utilize a boundary loss for positive samples (if $o$ is larger than a threshold value $\lambda$, the sample is viewed as positive sample) to promote localizing precise start and end points as follows:
\begin{equation}
    \mathcal{L}_{boundary} = \frac{1}{R_{pos}} \sum \mathcal{L}_1(\delta_s - \delta_s^{gt}) + \mathcal{L}_1(\delta_e - \delta_e^{gt}),
\end{equation}
where $R_{pos}$ is the number of positive samples, $\mathcal{L}_1$ denotes the smooth L1 function. The overall loss function can be formulated with a balanced parameter $\alpha$ as:
\begin{equation}
    \mathcal{L} = \mathcal{L}_{iou} + \alpha \mathcal{L}_{boundary}.
\end{equation}

\section{Experiments}
\subsection{Datasets and Evaluation Metric}
\noindent \textbf{Charades-STA.}
Charades-STA is a benchmark dataset for the video grounding task, which is built upon the Charades \cite{sigurdsson2016hollywood} dataset. It is collected for video action recognition and video captioning, and contains 6672 videos and involves 16128 video-query pairs.
Following previous work \cite{gao2017tall}, we utilize 12408 video-query pairs for training and 3720 pairs for testing.

\noindent \textbf{TACoS.}
TACoS is collected by \cite{regneri2013grounding} for video grounding and dense video captioning tasks. It consists of 127 videos on cooking activities with an average length of 4.79 minutes. In video grounding task, it contains 18818 video-query pairs. We follow the same split of the dataset as \cite{gao2017tall} for fair comparisons, which has 10146, 4589, and 4083 video-query pairs for training, validation, and testing respectively.

\noindent \textbf{Evaluation metric.}
We adopt “R@n, IoU=m” proposed by \cite{hu2016natural} as the evaluation metric, which calculates the IoU between the top-n retrieved video segments and the ground truth. It means the percentage of IoU greater than m.

\subsection{Implementation Details}
For appearance-aware object features, 
We utilize ResNet50 \cite{he2016deep} based Faster R-CNN \cite{ren2015faster} model pretrained on Visual Genome dataset \cite{krishna2016visual} to obtain appearance-aware object features, and extract its global feature from another ResNet50 pretrained on ImageNet \cite{deng2009imagenet}. The number $K$ of extracted objects is set to 20. For motion-aware object features, we define continuous 16 frames as a clip and each clip overlaps 8 frames with adjacent clips. We first extract clip-wise features from a pretrained C3D \cite{tran2015learning} or I3D \cite{carreira2017quo} model, and then apply RoIAlign \cite{he2017mask} on them to generate object features. 
Since some videos are overlong, we uniformly downsample frame- and clip-feature sequences to $T=256$.
As for sentence encoding, we utilize Glove \cite{pennington2014glove} to embed each word to 300 dimension features. The head size of multi-head self-attention is 8, and the hidden dimension of Bi-GRU is 512. The dimension $D$ is set to 1024, and the balance hyper-parameter $\alpha$ is set to 0.005. For segment proposals in grounding head, we have 800 samples for each video on both Charades-STA and TACoS datasets, and set $\lambda = 0.55$.
We train the whole model for 100 epochs with batch size of 16 and early stopping strategy.
Parameter optimization is performed by Adam optimizer with leaning rate $4\times 10^{-4}$ for Charades-STA and $3\times 10^{-4}$ for TACoS, and linear decay of learning rate and gradient clipping of 1.0. 

\begin{table*}[t!]
    \small
    \centering
    \begin{tabular}{c|c|cccc|c|cccc}
    \hline \hline
    \multirow{3}*{Method} & \multicolumn{5}{c|}{Charades-STA} & \multicolumn{5}{c}{TACoS} \\ \cline{2-11}
    ~ & \multirow{2}*{Feature} & R@1, & R@1, & R@5, & R@5, & \multirow{2}*{Feature} & R@1, & R@1, & R@5, & R@5, \\ 
    ~ & ~ & IoU=0.5 & IoU=0.7 & IoU=0.5 & IoU=0.7 & ~ & IoU=0.3 & IoU=0.5 & IoU=0.3 & IoU=0.5 \\ \hline
    CTRL & C3D & 23.63 & 8.89 & 58.92 & 29.57 & C3D & 18.32 & 13.30 & 36.69 & 25.42 \\
    QSPN & C3D & 35.60 & 15.80 & 79.40 & 45.50 & C3D & 20.15 & 15.32 & 36.72 & 25.30 \\
    CBP & C3D & 36.80 & 18.87 & 70.94 & 50.19 & C3D & 27.31 & 24.79 & 43.64 & 37.40 \\
    GDP & C3D & 39.47 & 18.49 & - & - & C3D & 24.14 & - & - & - \\
    VSLNet & I3D & 47.31 & 30.19 & - & - & C3D & 29.61 & 24.27 & - & - \\
    BPNet & I3D & 50.75 & 31.64 & - & - & C3D & 25.96 & 20.96 & - & - \\
    IVG-DCL & I3D & 50.24 & 32.88 & - & - & C3D & 38.84 & 29.07 & - & - \\
    DRN & I3D & 53.09 & 31.75 & 89.06 & 60.05 & C3D & - & 23.17 & - & 33.36 \\
    CBLN & I3D & 61.13 & 38.22 & 90.33 & 61.69 & C3D & 38.98 & 27.65 & 59.96 & 46.24 \\ \hline
    \multirow{2}*{\textbf{Ours}} & C3D+Object & 64.47 & 43.09 & 93.61 & 71.55 & C3D+Object & 46.33 & 35.74 & 63.97 & 53.18 \\
    ~ & I3D+Object & \textbf{66.43} & \textbf{44.80} & \textbf{95.57} & \textbf{73.26} & I3D+Object & \textbf{48.47} & \textbf{37.25} & \textbf{66.39} & \textbf{54.61} \\ \hline
    \end{tabular}
    \vspace{-6pt}
    \caption{Overall performance comparison among our method with proposal-based and proposal-free methods on the Charades-STA and TACoS datasets under the official train/test splits.}
    \vspace{-4pt}
    \label{tab:sota1}
\end{table*}

\begin{table*}[t!]
    \small
    \centering
    \begin{tabular}{c|c|cccc|c|cccc}
    \hline \hline
    \multirow{3}*{Method} & \multicolumn{5}{c|}{Charades-STA} & \multicolumn{5}{c}{TACoS} \\ \cline{2-11}
    ~ & \multirow{2}*{Feature} & R@1, & R@1, & R@5, & R@5, & \multirow{2}*{Feature} & R@1, & R@1, & R@5, & R@5, \\ 
    ~ & ~ & IoU=0.5 & IoU=0.7 & IoU=0.5 & IoU=0.7 & ~ & IoU=0.3 & IoU=0.5 & IoU=0.3 & IoU=0.5 \\ \hline
    MMRG & Object & 44.25 & - & 60.22 & - & Object & 57.83 & 39.28 & 78.38 & 56.34 \\ \hline
    \multirow{2}*{\textbf{Ours}} & C3D+Object & 48.63 & 34.25 & 65.79 & 41.10 & C3D+Object & 61.07 & 42.48 & 81.91 & 60.33 \\
    ~ & I3D+Object & \textbf{50.07} & \textbf{35.82} & \textbf{68.46} & \textbf{42.97} & I3D+Object & \textbf{63.69} & \textbf{44.37} & \textbf{85.28} & \textbf{62.14} \\ \hline
    \end{tabular}
    \vspace{-6pt}
    \caption{Comparison with detection-based method MMRG on Charades-STA and TACoS datasets under MMRG's train/test splits. We do not compare with \cite{zhang2020object,zhang2020does} since they address different tasks and datasets and are close source.}
    \label{tab:sota2}
    \vspace{-10pt}
\end{table*}


\subsection{Comparison with State-of-the-Arts}
\noindent \textbf{Compared methods.}
To demonstrate the effectiveness of our MARN, we compared it with several state-of-the-art methods: (1) Proposal-based: CTRL \cite{gao2017tall}, QSPN \cite{xu2019multilevel}, BPNet \cite{xiao2021boundary}, DRN \cite{zeng2020dense}, CBLN \cite{liu2021context}; (2) Proposal-free: CBP \cite{wang2019temporally}, GDP \cite{chenrethinking}, VSLNet \cite{zhang2020span}, IVG-DCL \cite{nan2021interventional}; (3) Detection-based: MMRG \cite{zeng2021multi}.

\noindent \textbf{Comparison on Charades-STA.}
We compare our MARN with the state-of-the-art proposal-based and proposal-free methods on the Charades-STA dataset in Table \ref{tab:sota1}, where we reach the highest results over all evaluation metrics. Particularly, our C3D+Object variant outperforms the best proposal-based method CBLN by 4.87\% and 9.86\% absolute improvement in terms of R@1, IoU=0.7 and R@5, IoU=0.7, respectively. Compared to the proposal-free method IVG-DCL, the C3D+Object model outperforms it by 14.23\% and 10.21\% in terms of R@1, IoU=0.5 and R@1, IoU=0.7, respectively. We also compare our model with the detection-based method MMRG in Table \ref{tab:sota2}. To make a fair comparison, we follow the same data splits for training/testing. It shows that our C3D+Object model brings a further improvement of 4.38\% and 5.57\% in terms of R@1, IoU=0.5 and R@5, IoU=0.5. We further utilize the I3D to present a new I3D+Object variant, which performs better than C3D+Object since I3D can obtain stronger features.

\noindent \textbf{Comparison on TACoS.}
Table \ref{tab:sota1} and \ref{tab:sota2} also report the grounding results on TACoS. Compared to CBLN, our C3D+Object model outperforms it by 7.35\%, 8.09\%, 4.01\%, and 6.94\% in terms of all metrics, respectively. Our model also outperforms IVG-DCL by a large margin. Compared to the detection-based method MMRG, our C3D+Object model brings the improvements of 3.20\% and 3.99\% in strict metrics of R@1, IoU=0.5 and R@5, IoU=0.5, respectively. The I3D+Object variant further achieves better results.

\begin{table}[t!]
\small
    \centering
    \begin{tabular}{l|cc}
    \hline \hline
    \multirow{3}*{Method} & \multicolumn{2}{c}{Charades-STA} \\ \cline{2-3}
     ~ & R@1, & R@5, \\
     ~ & IoU=0.7 & IoU=0.7 \\ \hline
     baseline & 34.76 & 63.09 \\ \hline
     + AB & 37.20 & 65.97 \\
     + AB\&ME & 38.91 & 67.64 \\
     + AB\&ME\&MB & 40.85 & 69.19 \\
     + AB\&ME\&MB\&MAA & \textbf{43.09} & \textbf{71.55} \\ \hline
    \end{tabular}
    \vspace{-6pt}
    \caption{Main ablation study on MARN under the official train/test splits. It investigates the appearance branch (AB), the motion encoder (ME), the motion branch (MB), and the motion-appearance associating module (MAA).}
    \label{tab:ablation1}
    \vspace{-15pt}
\end{table}

\subsection{Ablation Study}
In this section, we will perform in-depth ablation studies to evaluate the effectiveness of each component in our MARN on Charades-STA dataset. Since most previous works utilize C3D to extract features in this task and our C3D+object variant already achieves the state-of-the-art performance, we utilize the C3D+Object variant as our backbone here.

\noindent \textbf{Main ablation.}
We first perform main ablation studies to demonstrate the effectiveness of each component. To construct the baseline model, we utilize a general ResNet50 based Faster-RCNN for appearance-aware object extraction and another ResNet50 for frame-level global feature extraction, and do not encode any motion contexts. Instead of building the appearance branch, we employ a co-attention \cite{lu2016hierarchical} module to interact object-level cross-modal information and simply concatenate query-object features for semantic enhancement. We also utilize another co-attention module to capture object-relations, and a mean-pooling layer to fuse object features to represent frame-level features. We utilize the same grounding head in all ablation variants.
The performances of each variants are shown in Table \ref{tab:ablation1}, and we can observe the following conclusions: (1) It is worth noticing that the baseline model achieves better performance than all existing methods in Table \ref{tab:sota1}, demonstrating that the detection-based method is more effective in distinguishing the frames with high similarity. Our object-level features filter out redundant background information in frame-level features of previous works, thus leading to fine-grained activity understanding and more precise localization. 
(2) The proposed appearance branch (AB) can capture more fine-grained object relations, thus bringing a significant improvement. (3) The motion encoder (ME) and the corresponding motion branch (MB) can incorporate action-oriented contexts into appearance-based features for better understanding the activity. (4) Motion-appearance association module (MAA) further brings improvement, which proves the effectiveness of incorporating appearance and motion features for bi-directional enhancement.

\begin{table}[t!]
\small
    \centering
    \begin{tabular}{l|cc}
    \hline \hline
    \multirow{3}*{Method} & \multicolumn{2}{c}{Charades-STA} \\ \cline{2-3}
     ~ & R@1, & R@5, \\
     ~ & IoU=0.7 & IoU=0.7 \\ \hline
     MARN & \textbf{43.09} & \textbf{71.55} \\ \hline
     w/o global feature & 41.46 & 70.03 \\
     w/o position encoding & 40.73 & 68.69  \\ \hline
    \end{tabular}
    \vspace{-6pt}
    \caption{Ablation study on the video encoding.}
    \vspace{-2pt}
    \label{tab:ablation2}
\end{table}

\begin{table}[t!]
\small
    \centering
    \begin{tabular}{c|c|cc}
    \hline \hline
    \multirow{3}*{Module} & \multirow{3}*{Changes} & \multicolumn{2}{c}{Charades-STA} \\ \cline{3-4}
     ~ & ~ & R@1, & R@5, \\
     ~ & ~ & IoU=0.7 & IoU=0.7 \\ \hline
     \multirow{2}*{\tabincell{c}{Cross-modal\\Interaction }} & w/ attention & \textbf{43.09} & \textbf{71.55}  \\ 
     ~ & w/ concatenation & 41.62 & 70.17  \\ \hline
     \multirow{4}*{\tabincell{c}{Graph\\Network}} & w/ graph & \textbf{43.09} & \textbf{71.55}  \\ 
     ~ & w/o graph & 41.38 & 69.93 \\ \cline{2-4}
     ~ & layer=1 & \textbf{43.09} & \textbf{71.55} \\
     ~ & layer=2 & 42.74 & 71.26 \\ \hline
     \multirow{2}*{\tabincell{c}{Object-feature\\Fusion}} & w/ attention & \textbf{43.09} & \textbf{71.55}  \\ 
     ~ & w/ pooling & 41.81 & 70.44 \\ \hline
    \end{tabular}
    \vspace{-6pt}
    \caption{Ablation study on the reasoning branches.}
    \vspace{-2pt}
    \label{tab:ablation3}
\end{table}

\begin{table}[t!]
\small
    \centering
    \setlength{\tabcolsep}{1.2mm}{
    \begin{tabular}{cc|cc}
    \hline \hline
    \multirow{3}*{Motion-guided} & \multirow{3}*{Appearance-fused} & \multicolumn{2}{c}{Charades-STA} \\ \cline{3-4}
    ~ & ~ & R@1, & R@5, \\
    ~ & ~ & IoU=0.7 & IoU=0.7 \\ \hline
    $\times$ & $\times$ & 40.85 & 69.19 \\
    $\checkmark$ & $\times$ & 41.97 & 70.23 \\
    $\times$ & $\checkmark$ & 42.14 & 70.60 \\
    $\checkmark$ & $\checkmark$ & \textbf{43.09} & \textbf{71.55} \\ \hline
    \end{tabular}}
    \vspace{-6pt}
    \caption{Ablation study on the MAA module.}
    \label{tab:ablation4}
    \vspace{-15pt}
\end{table}

\noindent \textbf{Analysis on the video encoder.}
As shown in Table \ref{tab:ablation2}, we conduct the investigation on different video encoding. We find that the full model performances worse if we remove the global feature learning. It demonstrates that the frame-wise features help to better explore the non-local object information in the frame. Besides, it also presents the
effectiveness of the position encoding in identifying temporal semantic and improving the accuracy of temporal grounding.

\noindent \textbf{Analysis on the reasoning branches.}
We also conduct ablation study within the reasoning branches as shown in Table \ref{tab:ablation3}. For object-level cross-modal interaction, our attention based mechanism outperforms the mechanism of concatenating query-object features by 1.47\% and 1.38\%. For spatio-temporal graph network, replacing the graph model with simple co-attention model will reduce the performance, since the latter lacks temporal modeling. We can observe that our model achieves best result when the number of graph layer is set to 1, and more graph layers will result in over-smoothing problem \cite{li2018deeper}. For frame-level object-feature fusion, attention based mechanism performs better than the mean-pooling.

\noindent \textbf{Analysis on the associating module.}
As shown in Table \ref{tab:ablation4}, equipped with only
motion-guided appearance enhancement, there is a 1.12\% and 1.04\% point gain compared to the model w/o MAA. 
This is because the motion context contains implicit appearance information. 
Beside, when adopting appearance-fused motion enhancement alone, we achieve an improvement of 1.29\% and 1.41\%, since the appearance-aware objects can also contribute to motion modeling.
Applying both of them together, the performance boost is larger than using only one of them.

\begin{figure}[t]
\centering
\includegraphics[width=0.48\textwidth]{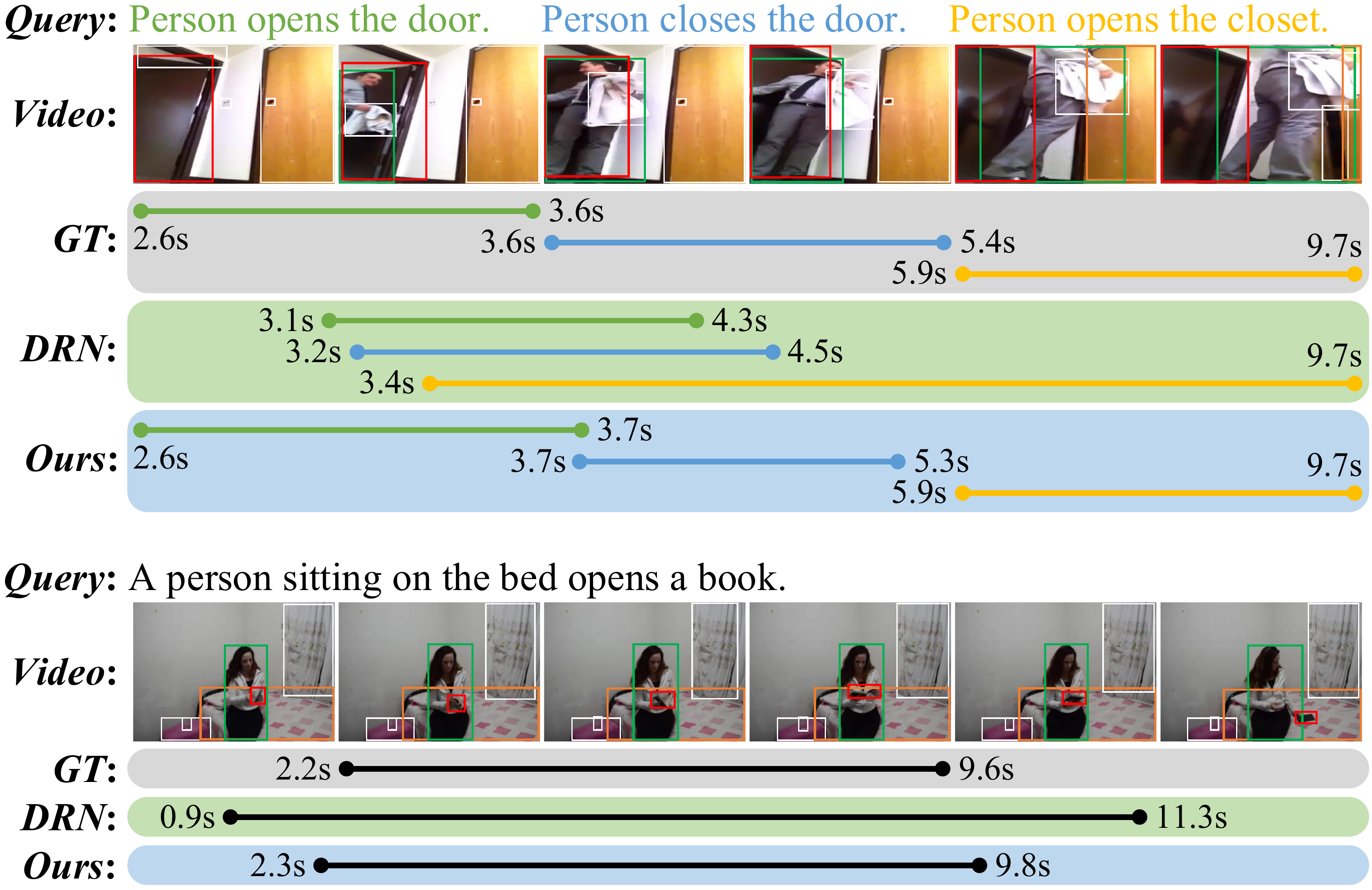}
\vspace{-14pt}
\caption{The qualitative results on Charades-STA dataset.}
\label{fig:result}
\vspace{-10pt}
\end{figure}

\subsection{Visualization}
We provide qualitative results in Figure \ref{fig:result}, where we choose DRN for comparison due to its open-source. Here, we only show a fixed number of bounding boxes, and color the best matching ones according to the attentive weights in Eq. (5) and (9).
For the first video, DRN relies on the frame-level video features, thus failing to distinguish the similar object ``door", ``closet", leading to worse grounding results. By contrast, our model utilizes a detection-based framework that easily captures the appearance differences between ``door" and ``closet". Besides, we also incorporate the object-level motion contexts into appearance features, providing more fine-grained details for action understanding.
For the second video,  our method captures the ``open" relations between ``person" and ``book", and performs more accurate grounding than DRN.

\section{Conclusion}
In this paper, we proposed a novel Motion-Appearance Reasoning Networks (MARN) for temporal sentence grounding, which incorporates both motion contexts and appearance features for better reasoning spatio-temporal semantic relations between objects. Through the developed motion and appearance branches, our MARN manages to mine both motion and appearance clues which matches the semantic of query, and then we devise an associating module to integrate the motion-appearance information for final grounding. Experimental results on two challenging datasets show the effectiveness of our proposed MARN. 
In the future, we will explore more robust query and video encoders (e.g. bert \cite{devlin2018bert} and X3D \cite{feichtenhofer2020x3d}) to further improve the grounding performance. 

\section{Acknowledgements} 
This work was supported in part by the National Natural Science Foundation of China (NSFC) under Grant No.61972448, and in part by the Shenzhen Basic Research (General Project) under Grant No.JCYJ20190806142601687.

\bibliography{reference.bib}

\end{document}